# Proprioceptive multistable mechanical metamaterial via soft capacitive sensors

Hugo de Souza Oliveira
*Cluster of Excellence livMatS@FIT*
*University of Freiburg*
79110 Freiburg, Germany
hugo.oliveira@livmats.uni-freiburg.de

Niloofar Saeedzadeh Khaanghah
*Faculty of Engineering*
*Free University of Bozen-Bolzano*
39100 Bozen-Bolzano, Italy
nsaeedzadehkhaanghah@.unibz.it

Martijn Oetelmans
*Faculty of Mechanical Engineering*
*Delft University of Technology*
2628CD, Delft, The Netherlands
p.m.oetelmans@student.tudelft.nl

Niko Münzenrieder
*Faculty of Engineering*
*Free University of Bozen-Bolzano*
39100 Bozen-Bolzano, Italy
niko.muenzenrieder@unibz.it

Edoardo Milana
*Department of Microsystems Engineering - IMTEK*
*University of Freiburg*
79110 Freiburg, Germany
edoardo.milana@imtek.uni-freiburg.de

*Abstract*—The technological transition from soft machines to soft robots necessarily passes through the integration of soft electronics and sensors. This allows for the establishment of feedback control systems while preserving the softness of the robot embodiment. Multistable mechanical metamaterials are excellent building blocks of soft machines, as their nonlinear response can be tuned by design to accomplish several functions. In this work, we present the integration of soft capacitive sensors in a multistable mechanical metamaterial, to enable proprioceptive sensing of state changes. The metamaterial is a periodic arrangement of 4 bistable unit cells. Each unit cell has an integrated capacitive sensor. Both the metastructure and the sensors are made of soft materials (TPU) and are 3D printed. Our preliminary results show that the capacitance variation of the sensors can be linked to state transitions of the metamaterial, by capturing the nonlinear deformation.

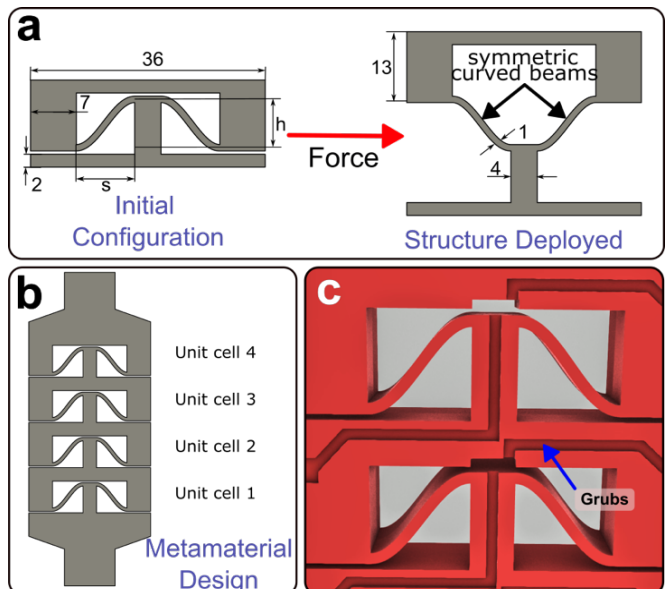

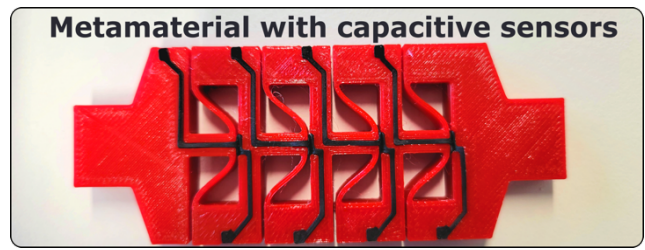

Fig. 1 – Metamaterial with integrated capacitive sensors for proprioception.

Fig. 2 – Design of the metamaterial structure. a) bistable unit cell utilized; b) metamaterial structure based on four unit cells. c) grubs to accommodate contacts and capacitive plates.

## I. Introduction

Over the last years, soft robotics has gained significant importance in sectors such as the food industry [1], minimally invasive surgery [2], or infrastructure safety [3], which implement automated processes and services that require delicate interactions with the environment. Soft robots embrace mechanical compliance to be safe, adaptive, and resilient, offering a mechanical alternative to torque or impedance control systems [4].

Recently, soft structures with nonlinear mechanical responses (e.g. balloons, slender beams, thin shells) have been integrated into soft robotic systems to augment their functionalities. Nonlinear responses, such as snap-through instabilities, can be leveraged to amplify displacement [5], to make jumps [6], to generate travelling waves [7] and self-oscillations [8], and even to play melodies on a piano using a single signal [9]. These functions are programmed by design and work as feedforward schemes.

However, autonomous systems require the implementation of feedback loops to truly self-regulate those functions. Indeed, like traditional robots, soft robots need sensors for control purposes and data collection, as well as electronic circuits to process those data and close the feedback loop between actuators and sensors. In this regard, the evolution of sensors and flexible electronics has been very important. Advances have enabled the creation of imperceptible, unobtrusive, lightweight, and transparent electronics [10], which starkly contrast with traditional bulky and rigid electronics. Integrating these soft electronics and sensors with soft robots is crucial for enhancing their applications, in terms of proprioception (i.e. self-awareness), as well as exteroception (i.e. environmental awareness).

Herein, we present a strategy to integrate proprioceptive sensing in a nonlinear soft structure for potential integration in soft robotic systems. We focus on a multistable mechanical metamaterial, a structure that can snap several times under tension in different stable states. The metamaterial structure comprises four bistable unit cells with integrated flexible conductive material, which work as capacitive sensors, allowing us to identify the sequence of state transitions of the metamaterial under tension. Fig. 1 shows the fabricated structure, with the capacitive structure embedded in the soft metamaterial.

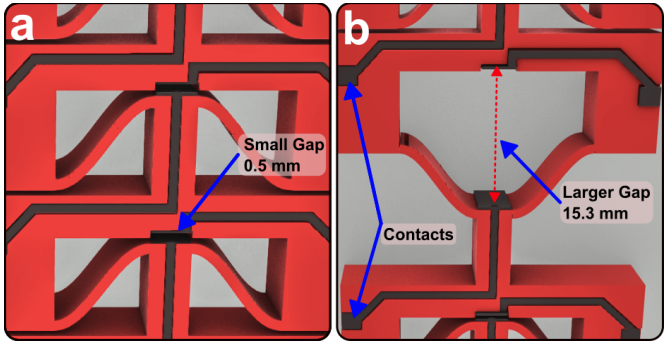

Fig. 3 – Integrating conductive materials. a) Closed unit cells with conductive plates and a small gap; b) Deployed unit cells with larger gap.

## II. DESIGN AND FABRICATION

The metamaterial structure is designed from a bistable unit-cell [11], which is composed of two symmetric curved beams that snap between two states when a tensile force is applied, as displayed in Fig. 2a. The curved beam $B$ is generated by the relation $B(s) = (h/2)(1 - \cos 2\pi(s/l))$, in which $s$ ranges from 0 to 9 $mm$, $h$ is 8 $mm$, and $l = 18\ mm$. The metamaterial has a thickness of 7 $mm$. Four unit cells are structured to enable multistability, as depicted in Fig. 2b. In the design, grubs are considered for accommodating the plates of the capacitor and the connections, as shown in Fig. 2c. The capacitor consists of two parallel conductive plates separated by an initial gap of 0.5 $mm$. When the unit cells are deformed in the other state, the distance between the plates increases to 15.3 $mm$, as shown in Fig. 3a and 3b. The dielectric materials of the capacitors are air and a thin vinyl layer to avoid electrical contact between the plates when the structure is not deformed. The flexible capacitive plates have a dimension 3.8 × 7.3 $mm$ and a thickness of 0.4 $mm$. The capacitive plates and the connections are 3D printed to match exactly the shape of the grubs on the metamaterial structure.

The metamaterial structure, the flexible capacitive plates, and connections are 3D printed with a Prusa MK4 3D printer. The structure is printed using a thermoplastic polyurethane (TPU) filament (NinjaTek Cheetah Shore 95A) with 80% of infill and a grid fill pattern, while the conductive parts are printed using carbon-filled TPU conductive filament (Recreus Filaflex Conductive Shore 92A) with 100% infill. The capacitors and the connections are then placed in the grubs and fused with the metamaterial structure using a soldering process.

## III. RESULTS AND DISCUSSION

The multistable metamaterial is tested in a uniaxial tensile test machine (ZwickRoell Z010) with controlled displacement of 1 $mm/s$ to evaluate its nonlinear characteristic and to sense the state transitions (deployment) sequence of the unit cells. The signals from the capacitive sensors are acquired using a Texas Instruments FDC2214EVM module with 28-bit capacitance-to-digital converter evaluation module. Each capacitor is connected to one of the four channels of the board. Although the board does not directly measure the capacitance, it detects variations in capacitance by forming an inductive capacitive (LC) circuit, where changes in capacitance alter the resonant frequency of the circuit. The board excites the LC circuit, measures the oscillation frequency, and converts this frequency shift into a digital value. The digital output corresponds to the detected change in capacitance.

### A. Mechanical properties

Fig. 4 shows the deployment sequence of the unit cells. Although the unit cells are fabricated using identical geometrical parameters and printing settings, imperfections occur and dictate which unit cell requires lower force to undergo snapping before the other unit cells. When the structure is pulled, Unit Cell 2 (U2) is the first to change state, followed by U3, U1, and finally U4.

Fig. 5a shows 10 loading cycles, exhibiting the consistent repeatability of the nonlinear mechanical response. Fig. 5b depicts a single tensile loading of the structure, clearly showing the four different force peaks that correspond to the deployment of each unit cell. The "softness" of the structure is reflected by the low elastic forces generated (up to 7 $N$).

### B. Electrical properties

Fig. 5a also shows the variation of the capacitance over 10 cycles, demonstrating also good stability and repeatability over time for the four sensors. Fig. 5c shows the normalized variation of the capacitance of the sensors during the tensile load. The signals are affected by several nonlinearities induced by several causes. First, all unit cells undergo a certain amount of deformation, causing a capacitance variation in all of them. Moreover, the mutual capacitance of the other capacitors has an influence as well on the signals. This effect can be added up to the interaction of the electrical field of capacitors and the electrical connections, which also contributes to the capacitance variations, therefore to variations in all the other capacitances.

Nevertheless, the variation in the capacitance of the sensor that is attached to the unit cell under deployment is higher, due to the larger displacement between plates. This effect can be better visualized through the first derivative of the normalized sensor signals, plotted in Fig. 5c, where the peak represents the highest variation. This peak consistently occurs during the unit cell deployment, highlighted by the coloured background areas in Fig. 5c. This analysis allows the detection of the deployment sequence. By simple calculations, considering the geometry of the capacitive plates, the capacitance of each sensor when the unit cell is opened and closed are 0.01 $pF$ and 0.50 $pF$, respectively.

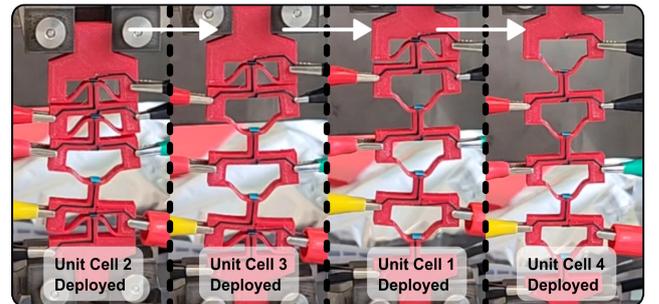

Fig. 4 – Deployment sequence of the unit cell in the tensile test.

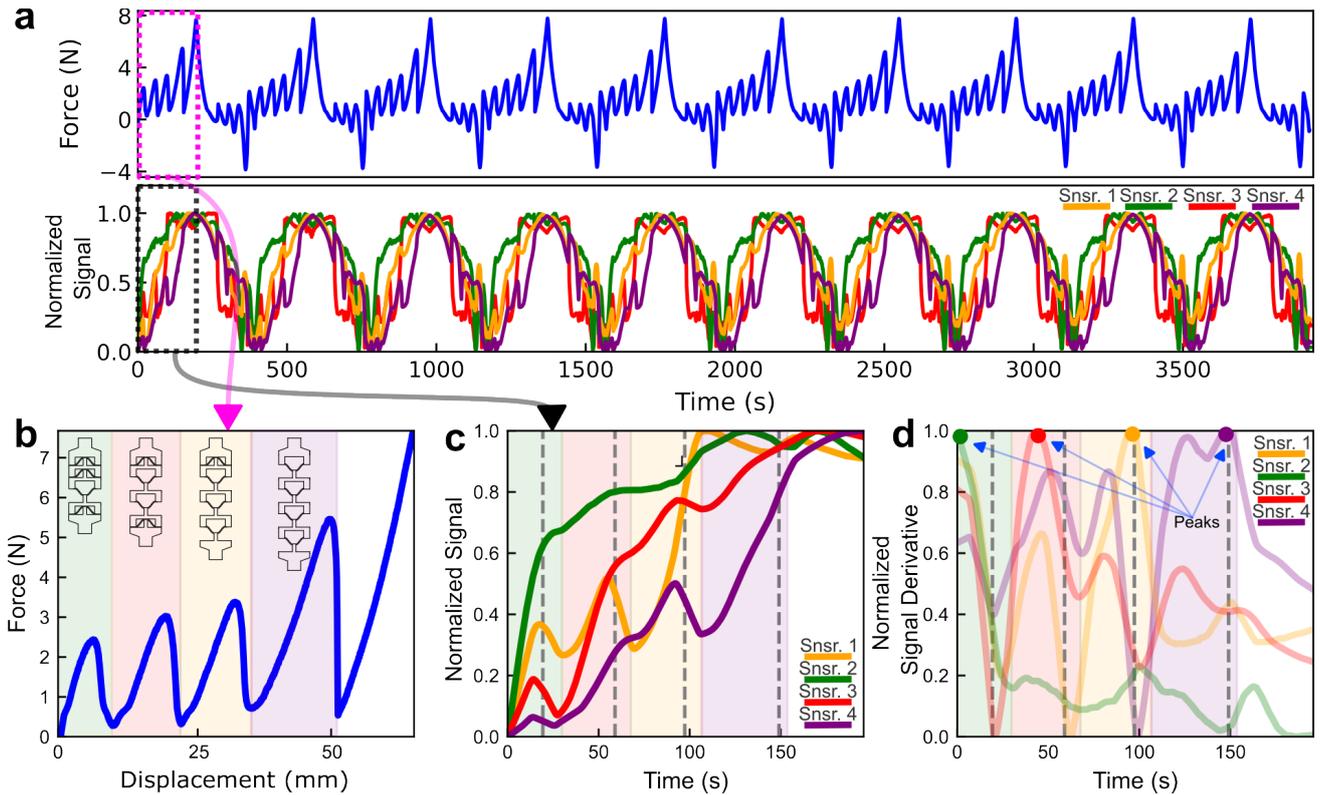

Fig. 5 – Mechanical and electrical properties. a) tensile and compression cycles of the metamaterial structure; b) nonlinear force-displacement curve of the metamaterial under a single tension load; c) normalized response from the sensor to the cycling tests; d) first derivative of the normalized response from the sensor.

## *C. Proprioceptive properties*

The states transition of the mechanical metamaterial is captured by the capacitance variation of the sensor. This variation is caused by the increased distance between the conductive plates during the deformation. Thus, the sensors integration provides the metamaterial with a proprioceptive ability, i.e. knowledge about the sequence of state transitions. Since there is no obvious way to determine the snapping sequence of the unit cells in multistable metamaterials, the integration of simple soft 3D printed capacitive sensors offers an attractive way to give metamaterials such proprioceptive capabilities.

Moreover, in real-world applications, snapping sequences may vary from the one in quasi-static conditions. External forces acting locally on unit cells, dynamic loading, fatigue effects, are all unpredictable effects that affect the mechanical properties of the metamaterial and, thus, modify the snapping force peaks. With the integration of proprioceptive monitoring, it would be possible for the metamaterial to know its sequence of state transitions, making it resilient to the external conditions that can modify the quasi-static sequence.

## CONCLUSION

Integrating soft capacitive sensors in multistable mechanical metamaterials is a promising approach to provide soft structures with proprioception. Our preliminary results demonstrate that this sensing approach is suitable to detect the state transitions sequence that occur in multistable metamaterials when deformed. In particular, we tested a metamaterial composed of 4 bistable unit cells integrated with 4 capacitive sensors. The system we proposed is entirely soft and 3D printable using commercially available flexible filaments based on TPU. Future work will focus on the improvement of the capacitive sensor design to reduce the influence of mutual capacitance and electrical connection. Eventually, we will integrate our sensing metamaterials in the design of nonlinear soft actuators, augmenting the capabilities of soft robotic systems and at the same time preserving their soft material embodiment.

## ACKNOWLEDGMENT

This work was funded by the Deutsche Forschungsgemein- schaft (DFG, German Research Foundation) under Germany's Excellence Strategy – EXC-2193/1 – 390951807 and under Walter Benjamin Programme – MI 3063/1-1 – 505075715.

## REFERENCES

[1] F. Hartmann, M. Baumgartner, and M. Kaltenbrunner, "Becoming Sustainable, The New Frontier in Soft Robotics," *Adv. Mater.*, vol. 33, no. 19, p. 2004413, 2021, doi: 10.1002/adma.202004413.
[2] M. Cianchetti, C. Laschi, A. Menciassi, and P. Dario, "Biomedical applications of soft robotics," *Nat. Rev. Mater.*, vol. 3, no. 6, pp. 143–153, Jun. 2018, doi: 10.1038/s41578-018-0022-y.
[3] E. Milana, "Soft robotics for infrastructure protection," *Front. Robot. AI*, vol. 9, Nov. 2022, doi: 10.3389/frobt.2022.1026891.
[4] C. D. Santina, M. G. Catalano, and A. Bicchi, "Soft Robots," pp. 1–15, 2020, doi: 10/gr8cxr.
[5] J. T. B. Overvelde, T. Kloek, J. J. A. D'haen, and K. Bertoldi, "Amplifying the response of soft actuators by harnessing snap-through instabilities," *Proc. Natl. Acad. Sci.*, vol. 112, no. 35, pp. 10863–10868, Sep. 2015, doi: 10.1073/pnas.1504947112.
[6] B. Gorissen, D. Melancon, N. Vasios, M. Torbati, and K. Bertoldi, "Inflatable soft jumper inspired by shell snapping," *Sci. Robot.*, vol. 5, no. 42, p. eabb1967, May 2020, doi: 10.1126/scirobotics.abb1967.
[7] E. Milana, B. Van Raemdonck, A. S. Casla, M. De Volder, D. Reynaerts, and B. Gorissen, "Morphological Control of Cilia-Inspired Asymmetric

Movements Using Nonlinear Soft Inflatable Actuators," *Front. Robot. AI*, vol. 8, Jan. 2022, doi: 10.3389/frobt.2021.788067.

[8] L. C. van Laake, J. de Vries, S. Malek Kani, and J. T. B. Overvelde, "A fluidic relaxation oscillator for reprogrammable sequential actuation in soft robots," *Matter*, vol. 5, no. 9, pp. 2898–2917, Sep. 2022, doi: 10.1016/j.matt.2022.06.002.

[9] B. Van Raemdonck, E. Milana, M. De Volder, D. Reynaerts, and B. Gorissen, "Nonlinear Inflatable Actuators for Distributed Control in Soft Robots," *Adv. Mater.*, vol. 35, no. 35, p. 2301487, 2023, doi: 10.1002/adma.202301487.

[10] F. Catania, H. de S. Oliveira, P. Lugoda, G. Cantarella, and N. Münzenrieder, "Thin-film electronics on active substrates: review of materials, technologies and applications," *J. Phys. Appl. Phys.*, vol. 55, no. 32, p. 323002, 2022, doi: 10/gr8cwk.

[11] H. Yang and L. Ma, "Multi-stable mechanical metamaterials by elastic buckling instability," *J. Mater. Sci.*, vol. 54, no. 4, pp. 3509–3526, Feb. 2019, doi: 10.1007/s10853-018-3065-y.